%% file: main.tex
\pdfoutput=1

\documentclass[11pt]{article}

\usepackage[]{acl}

\usepackage{times}
\usepackage{latexsym}

\usepackage[T1]{fontenc}

\usepackage[utf8]{inputenc}

\usepackage{amssymb}
\usepackage{bbm}
\usepackage{bm}
\usepackage{titlesec}
\usepackage{multirow}
\usepackage{mathtools}
\usepackage{colortbl}
\usepackage{float}
\usepackage{xcolor}
\usepackage{adjustbox}
\usepackage{enumitem}
\usepackage{arydshln}
\usepackage[capitalize]{cleveref}

\usepackage{booktabs}
\usepackage[font=small,labelfont=bf,tableposition=top]{caption}
\DeclareCaptionLabelFormat{andtable}{#1~#2  \&  \tablename~\thetable}

\usepackage{microtype}

\definecolor{RoseQuartzBg}{HTML}{F7CAC9}
\definecolor{RoseQuartz}{HTML}{F5A798}
\definecolor{Serenity}{HTML}{92A8D1}
\definecolor{OrangeRed}{rgb}{1.0, 0.27, 0.0}
\definecolor{Red}{rgb}{1.0, 0.0, 0.0}
\definecolor{Turquoise}{HTML}{0F4C81}
\usepackage{xparse}
\NewDocumentCommand{\lifu}{ mO{} }{\textcolor{OrangeRed}{\textsuperscript{\textit{Lifu}}\textsf{\textbf{\small[#1]}}}}
\NewDocumentCommand{\ken}{ mO{} }{\textcolor{purple}{\textsuperscript{\textit{Ken}}\textsf{\textbf{\small[#1]}}}}
\NewDocumentCommand{\zhe}{ mO{} }{\textcolor{blue}{\textsuperscript{\textit{Zhe}}\textsf{\textbf{\small[#1]}}}}

\definecolor{celadon}{rgb}{0.67, 0.88, 0.69}

\usepackage{tikz}
\newcommand*\circled[1]{\tikz[baseline=(char.base)]{
            \node[shape=circle,draw,inner sep=0.4pt] (char) {#1};}}

\title{PLANET: Dynamic Content Planning in Autoregressive Transformers for Long-form Text Generation}

\author{Zhe Hu$^{1}$,  Hou Pong Chan$^{2}$,  Jiachen Liu$^{1}$, Xinyan Xiao$^{1}$, Hua Wu$^{1}$, and Lifu Huang$^{3}$
\\
  $^{1}$Baidu Inc 
  $^{2}$Faculty of Science and Technology, University of Macau
  $^{3}$Virginia Tech
 \\
  $^{1}${\tt \{huzhe01,liujiachen,xiaoxinyan,wu\_hua\}@baidu.com} \\
  $^{2}${\tt hpchan@um.edu.mo}, 
  $^{3}${\tt lifuh@vt.edu}
  }

\begin{document}
\maketitle

\begin{abstract}
\input{abstract}
\end{abstract}

\section{Introduction}
\input{intro}

\section{Related Work}
\input{related}

\section{Our \textsc{PLANET} Framework}
\input{method}

\section{Experimental Setups}
\input{experiments}

\section{Results and Analysis}

\input{results}

\section{Conclusion}
\input{conclusion}

\section*{Acknowledgements}
We thank the anonymous reviewers, area chair, and senior area chairs for their constructive suggestions on our work. 
We also thank Xinyu Hua for the helpful discussions. 
Hou Pong Chan was supported by the Science and Technology Development Fund, Macau SAR (Grant No. 0101/2019/A2), and the Multi-year Research Grant from the University of Macau (Grant No. MYRG2020-00054-FST). Lifu Huang also thanks the support from the Amazon Research Awards.

\section*{Ethics Statement}
\input{ethics}

\bibliography{anthology,custom}
\bibliographystyle{acl_natbib}

\appendix
\section{Experiment Details}
\input{supplementary/details}

\section{Details for Human Evaluation}
\label{sec:human_eval_guideline}
\input{supplementary/human_eval}

\section{Discussions on Limitations and Future Directions}
\input{supplementary/limitations}

\section{Additional Sample Outputs}
\label{sec:additional_samples}
\input{supplementary/samples}

\end{document}

%% file: abstract.tex
Despite recent progress of pre-trained language models on generating fluent text, existing methods still suffer from  incoherence problems in long-form text generation tasks that require proper content control and planning to form a coherent high-level logical flow.
In this work, we propose PLANET, a novel generation framework leveraging autoregressive self-attention mechanism to conduct content planning and surface realization dynamically. 
To guide the generation of output sentences, our framework enriches the Transformer decoder with latent representations to maintain sentence-level semantic plans grounded by bag-of-words. 
Moreover, we introduce a new coherence-based contrastive learning objective to further improve the coherence of output. 
Extensive experiments are conducted on two challenging long-form text generation tasks including counter-argument generation and opinion article generation. Both automatic and human evaluations show that our method significantly outperforms strong baselines and generates more coherent texts with richer contents. 

%% file: intro.tex
Neural sequence-to-sequence (seq2seq) models are dominant methods for text generation nowadays, which are trained to maximize the log-likelihood over targets in an end-to-end fashion~\cite{cho-etal-2014-learning}. Recently,  pre-trained methods such as GPT-2~\cite{radford2019language} and BART~\cite{lewis-etal-2020-bart} have achieved promising results by leveraging large-scale data.
While these models can generate fluent results, they still fall short of producing coherent long-form texts with multiple sentences~\cite{dou2021scarecrow}.

\begin{figure}[t]
    \centering
    \includegraphics[width=\columnwidth,trim=0 0 0cm 0, clip]{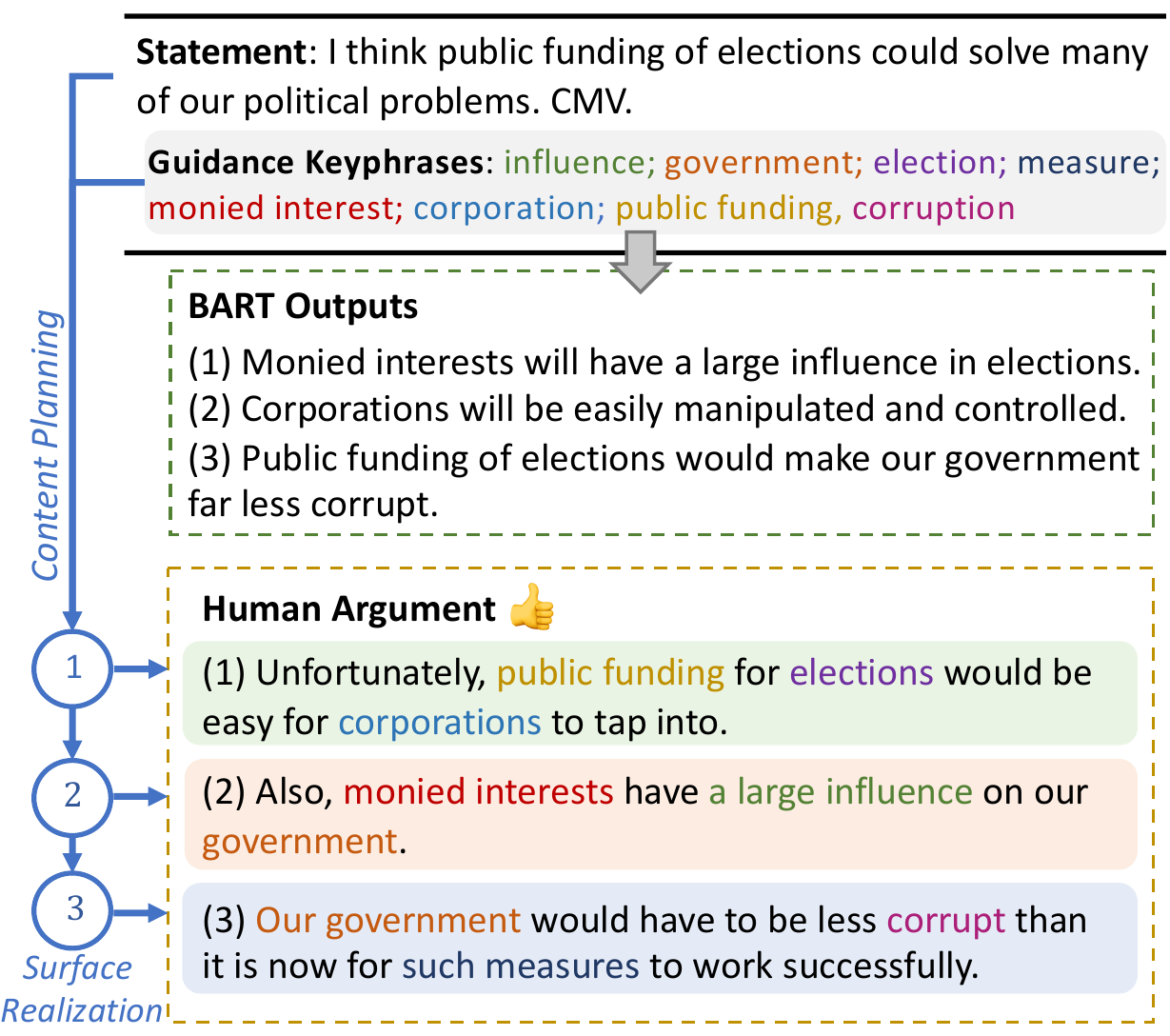}
    \captionof{figure}{ 
    Sample counter-arguments on Reddit ChangeMyView.
    Given a statement and a set of unordered keyphrases as guidance talking points,
    BART generates an incoherent output. In contrast, human writer conducts content planning and keyphrase selection for each sentence to form a coherent counter-argument.
    }
    \vspace{-3mm}
    \label{fig:sample_intro}
\end{figure}

Long text generation, especially opinion generation, usually requires the model to (1) conduct proper content 
selection and ordering 
(i.e., ``\textit{what to say and when to say it}'') to form a coherent high-level logical flow, and (2) appropriately reflect the text plans into final outputs (i.e.,``\textit{how to say it}'').
We present an example of counter-argument generation in Figure~\ref{fig:sample_intro}: given a statement on a controversial topic and a set of keyphrases as guidance talking points, the task aims to produce an argument with a different stance to refute the statement~\cite{hua-etal-2019-argument-generation}.
Human writer assigns keyphrases for each sentence to form a coherent logical flow (e.g., ``\textit{corporations easily tap into public funding}'' $\rightarrow$ "\textit{they also have large influence on government}" $\rightarrow$ "\textit{the current government is still corrupt}") and produces the final counter-argument that "\textit{public funding won't solve the election problems}".
In contrast, although BART learns to include keyphrases and generate an argument relevant to the statement, it suffers from incoherence issues such as incorrect usage of keyphrases (not ``\textit{corporations}'' but `\textit{election}'' that ``\textit{be manipulated and controlled}'') and wrong stance (``\textit{public funding would make government less corrupt}''),
and fails to maintain smooth transitions between sentences (e.g., sentence 2 and 3 are unrelated) and form a coherent text.

\begin{figure}[t]
    \centering
    \includegraphics[scale=0.54]{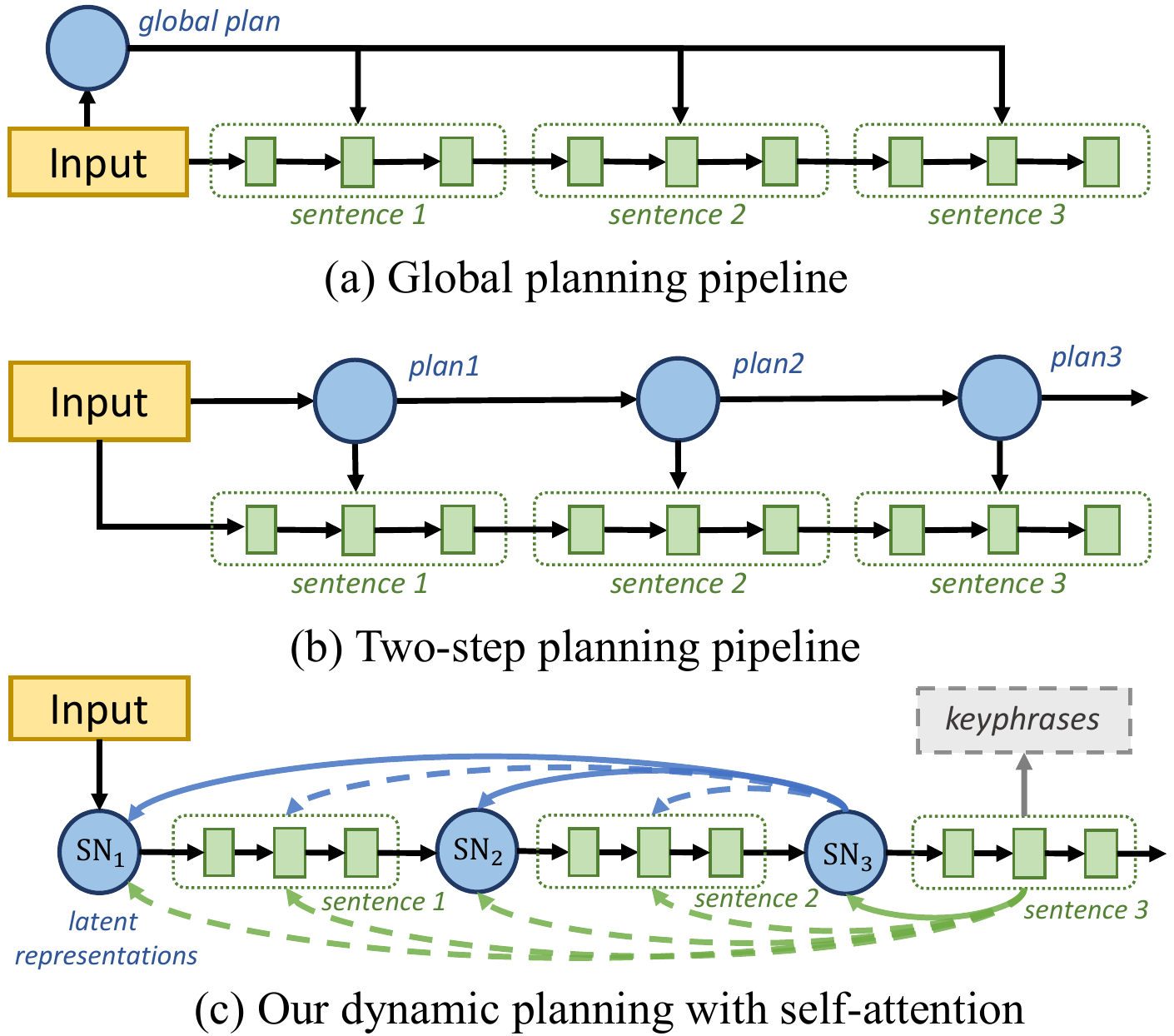}
    \captionof{figure}{ 
    Comparison of different content planning. For (c), the blue arrow denotes the attention flows for latent representations and the green one for target words. The attention of tokens within the same sentence is omitted. We highlight the attention flows related to the content planning with solid lines for sentence 3. Best viewed in color.
    }
    \vspace{-2mm}
    \label{fig:plan_type}
\end{figure}

To solve the above defects, various text planning methods were proposed to improve the coherence of the generated text. 
The first type of methods~\cite{kang-hovy-2020-plan,fu2020paraphrase,kong-etal-2021-stylized} leverage a latent variable as a global plan to guide the generation process, as illustrated in Figure~\ref{fig:plan_type}~(a). 
However, these methods do not consider fine-grained sentence-level planning. 
The second line of methods~\cite{hua-wang-2020-pair,goldfarb-tarrant-etal-2020-content} first produce sentence-level content plans, and then pass content plans to a surface realization module to generate the output words, as shown in Figure~\ref{fig:plan_type}~(b).  
Nevertheless, the planning and surface realization components are disjointed and may lead to cascading errors~\cite{hua-etal-2021-dyploc}. 
In this work, we propose \textbf{\textsc{PLANET}}, a novel text generation framework that dynamically performs content \underline{pl}anning and surf\underline{a}ce realizatio\underline{n} in autoregressiv\underline{e} \underline{T}ransformers.
As shown in Figure~\ref{fig:plan_type}~(c), for each target sentence, an autoregressive decoder first performs dynamic content planning by producing a latent representation ($\text{SN}_{j}$) as a semantic guidance, and then generates the sentence words. 
Both the content planning and surface realization are achieved dynamically by the autoregressive self-attention in a unified way: to generate a sentence (e.g., sentence $3$), the latent representation ($\text{SN}_{3}$) attends the previous latent representations ($\text{SN}_{1,2}$, solid blue arrows) and previous context (sentence $1$ and $2$, dashed blue arrows) to plan its overall semantic content; 
Then, each output position in the sentence attends the corresponding latent representation ($\text{SN}_{3}$, solid green arrow) and the previous words (dashed green arrows), and optionally select keyphrases 
(gray arrow) to decide the exact wording.
To supervise the latent representations, we further introduce a sentence-level bag-of-words prediction auxiliary task to provide supervision signals of the lexical semantics of the corresponding sentence. 
In this way, our framework can be trained end-to-end and easily applied to pre-trained autoregressive Transformers. 

Furthermore, to empower our model to distinguish coherent and incoherent targets and generate more coherent outputs,
we propose a novel coherence-based contrastive learning objective with different strategies to construct negative samples. We evaluate our model on two long-form opinion generation tasks: (1) counter-argument generation with Reddit/ChangeMyView dataset, and (2) opinion article generation from the New York Times Opinion corpus. Automatic evaluations show that our proposed method significantly outperforms strong baselines and generates more coherent texts with richer contents. Human evaluations further indicate that our model can properly leverage guidance keyphrases and generate better results on both datasets. 


The overall contributions of our work are:

\begin{itemize}[nolistsep,wide]
\item A unified framework that dynamically conducts content planning and surface realization by leveraging 
the autoregressive self-attention, with a novel sentence-level bag-of-words auxiliary task to guide the semantic content of each sentence;
\item A new coherence-based contrastive learning method with different negative sample construction strategies to improve the coherence of outputs;
\item Our approach outperforms strong baselines for both automatic and human evaluations on two challenging long-form text generation tasks.
\end{itemize}

%% file: related.tex
\noindent\textbf{Text Planning for Neural Generation.}
Traditional text generation pipeline leverages text planning component to decide on the high-level structures~\cite{mckeown1985discourse,reiter1997building,hovy1990pragmatics,CARENINI2006925}. 
Earlier work incorporates text planning into neural seq2seq structures by introducing hierarchical decoders~\cite{yao2019plan,moryossef-etal-2019-step,shen-etal-2019-towards}. However, these methods are hard to be applied to pre-trained models because of the modifications of model architecture. 
Several studies design separate modules for text planning and surface realization~\cite{hua-wang-2020-pair,tan-etal-2021-progressive,goldfarb-tarrant-etal-2020-content}, which lead to a disconnection of the two components and often produce undesired outputs~\cite{castro-ferreira-etal-2019-neural}. 
Recently, \citet{rashkin-etal-2020-plotmachines}
present a memory-based model to keep track of the content usage and generate paragraphs recurrently. Nevertheless, they do not consider sentence-level text planning which is critical to maintain high-level logical flow for opinion text generation. 
\citet{hua-etal-2021-dyploc} propose a mixed language model to perform content selection and ordering. However,
they encode multiple content items separately and do not fully consider the interactions among content items.
In contrast to these prior studies, our model conducts sentence-level text planning and surface realization dynamically by introducing high-level latent representations for target sentences, and can be incorporated into pre-trained autoregressive Transformers. 

\smallskip
\noindent\textbf{Coherent Long-form Text Generation.} 
Recent work tackles this problem on the tasks including story generation~\cite{fan-etal-2019-strategies,xu-etal-2020-megatron}, paragraph completion~\cite{kang-hovy-2020-plan},
text infilling~\cite{huang-etal-2020-inset},
long-form conversation~\cite{xu2021beyond} and news article generation~\cite{rashkin-etal-2020-plotmachines,tan-etal-2021-progressive}. To solve the incoherence issue, one type of work adopts the plan-then-generate strategy as discussed above. Some work also incorporates discourse and structured information into generation process to improve output coherence~\cite{jiang-etal-2021-enriching,ji-huang-2021-discodvt,bosselut-etal-2018-discourse}. Recently, \citet{guan-etal-2021-long} propose two auxiliary objectives of similarity prediction and order discrimination to improve coherence. 
In this work, we focus on long-form opinion text generation which requires an appropriate combination of credible talking points with rigorous reasoning~\cite{hua-etal-2019-argument-generation}, and apply dynamic content planning with a coherence-based contrastive objective to improve output coherence.

\smallskip
\noindent\textbf{Controllable Text Generation.}
Our work is closely related to controllable generation~\cite{prabhumoye-etal-2020-exploring}.
In this regard, typical studies manipulate sentiments~\cite{hu2017toward}, style~\cite{gao-etal-2019-structuring,du-ji-2021-sidecontrol-controlled,hu2021controllable}, syntax~\cite{chen-etal-2019-controllable}, and keywords~\cite{keskar2019ctrl,he2020ctrlsum,wu2020controllable} to steer the generation process.
We use topical keyphrases as guidance talking points and require the model to properly organize and reflect keyphrases for long-form opinion text generation.

%% file: method.tex
\begin{figure*}[t]
    \centering
    \includegraphics[scale=0.73]{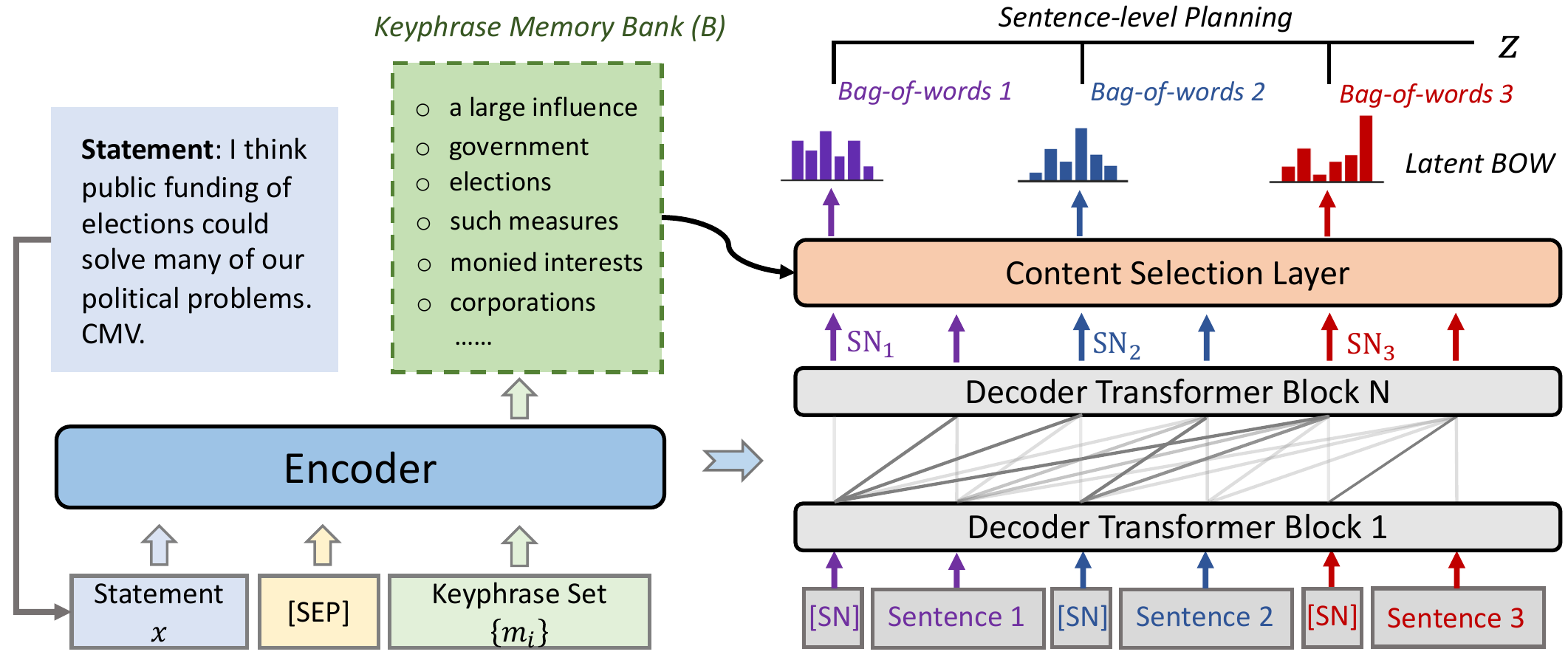}
    \captionof{figure}{ Overview of our framework. The encoder takes as input a statement and a set of keyphrases, and generates a keyphrase memory bank $\mathcal B$. 
    The decoder conducts content planning and surface realization dynamically by the autoregressive self-attention to produce a coherent output. Meanwhile, the latent representations (SN) predict bag-of-words as global semantic plans and guide the surface realization of each target sentence. We highlight attention flows related to the content planning.
    }
    \label{fig:decoder}
\end{figure*}

\subsection{Framework Overview}
\noindent\textbf{Task Description.}
We follow the previous work~\cite{hua-wang-2020-pair} and model the long-form opinion generation task by considering the input of (1) a statement $\bm x$ which can be a proposition for argument generation or a title for opinion-article generation, and (2) a set of unordered keyphrases $\bm m= \{\bm m_i\}$ related to the statement, serving as topical guidance signal. The output $\bm y$ is an opinion text consisting of multiple sentences and properly reflects the keyphrases in a coherent way.

Our framework is based on the seq2seq structure, and we adopt BART~\cite{lewis-etal-2020-bart} as the base model.
\footnote{Our method can be also applied to other autoregressive pre-trained language models.}
The overall framework is shown in Figure~\ref{fig:decoder}.
The bi-directional encoder first encodes the statement and keyphrases, and the decoder then generates the output in an autoregressive manner:

{\fontsize{10}{11}\selectfont
\vspace{-3mm}
\begin{align}
&{\bm{\hat{y}} = \text{argmax}\prod_{t=1}^{n}P({y_t|\bm{y}_{1:t-1},\bm x, \bm m)}
}\text{,}
\label{eq:prior}
\end{align}
\vspace{-3mm}
}

\noindent where $n$ is the number of target words. The statement and keyphrases are concatenated, with a segmenter inserted between adjacent keyphrases to indicate the keyphrase boundary.



\begin{figure}[t]
    \centering
    \includegraphics[scale=0.63]{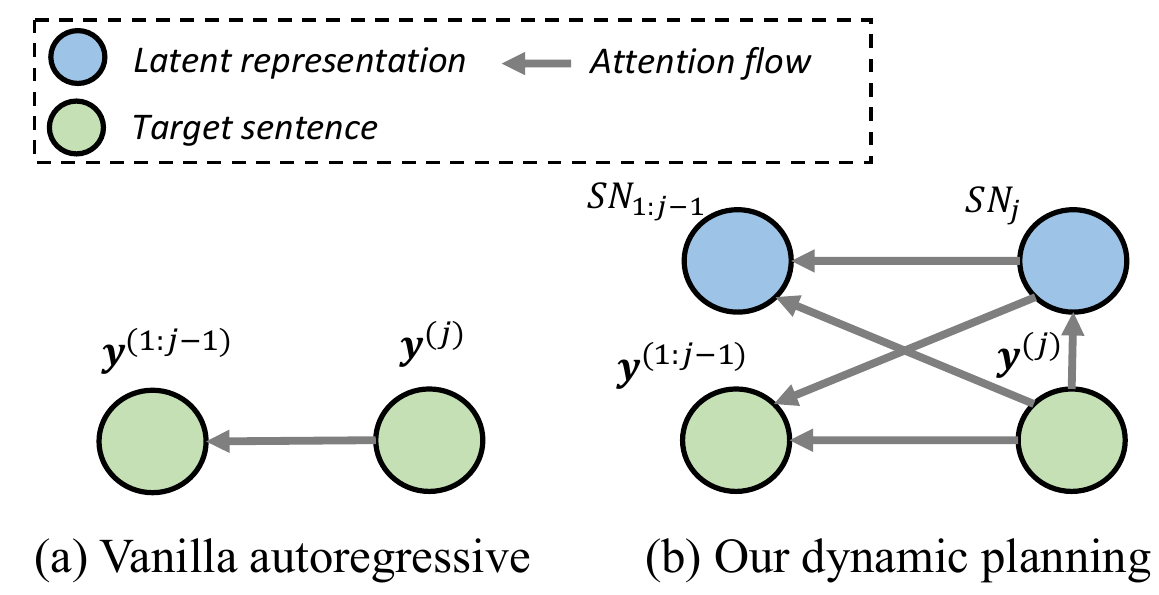}
    \captionof{figure}{ 
    Attention flow of our dynamic planning and surface realization. $\bm{y}^{(j)}$ represents the words of the $j$-th sentence.
    }
    \label{fig:our_flow}
\end{figure}


We conduct content planning and surface realization dynamically by leveraging the autoregressive self-attention mechanism. For each target sentence, we introduce a latent representation $\text{SN}$ to represent its global semantic information and guide surface realization (\S~\ref{subsec:latent}), then the sentence words attend the latent representation and dynamically select keyphrases (\S~\ref{subsec:content_selection}). After that, a sentence-level bag-of-words planning is introduced to enhance the latent representations (\S~\ref{subsec:bow}). Finally, we devise a contrastive learning (CL) objective to further improve the coherence of the output text (\S~\ref{subsec:cl}). 

\subsection{Latent Representation Learning}
\label{subsec:latent}
We introduce a latent representation for each target sentence to represent the overall semantic information and guide the generation of the sentence words. 
In particular, we insert a special token [SN] before every target sentence, and regard the hidden states of the decoder at the positions corresponding to [SN] as the latent representations of the target sentences. This has been shown effective by previous work~\cite{guan-etal-2021-long,li-etal-2021-conversations}.

The workflow of our dynamic planning and realization is shown in Figure~\ref{fig:our_flow}. 
For the vanilla autoregressive decoder, the generation of each token only depends on the previously generated tokens. 
In our framework, when producing the $j$-th output sentence $\bm{y}^{(j)}$, the latent representation $\text{SN}_j$ is first obtained by attending the previous latent representations $\text{SN}_{1:j-1}$ and words in previous sentences $\bm{y}^{(1:j-1)}$. 
Then for sentence-level surface realization, each token in the current sentence $\bm{y}^{(j)}$ attends the previously generated words and latent representations $\text{SN}_{1:j-1}$, as well as the current latent representation $\text{SN}_{j}$ as the guidance. 
A unique advantage of such modeling is that \textit{the content planning and surface realization can be performed simultaneously and incorporated into any pre-trained autoregressive language models, further optimized in an end-to-end fashion}.

\subsection{Content Selection}
\label{subsec:content_selection}
Based on the guidance of latent representations, each sentence word conducts content selection by incorporating keyphrases into decoder hidden states to decide which keyphrases to be reflected during generation.
We first feed the keyphrases to the encoder to obtain hidden representations. We then construct a keyphrase memory bank $\mathcal B$ by gathering the top layer representations of the segment tokens (each keyphrase is represented by the segment token before it). 
After that, a content selection layer retrieves keyphrase information from the keyphrase bank and integrates the selected information into the decoding process. 




\smallskip
\noindent \textbf{Content Selection Layer.}
At each decoding step $t$, the top layer representation of the Transformer decoder $\bm{h_t}$ attends the keyphrase memory bank via multi-head attention:

{\fontsize{10}{11}\selectfont
\vspace{-3mm}
\begin{align}
& \bm{c_t} = \textsc{MH-Attention}( 
\bm h_t, \mathcal B, \mathcal B)\text{,}
\end{align}
\vspace{-4mm}
}

\noindent where $\bm{c_t}$ is a context vector that embeds the selected keyphrase information, $\bm{h_t}$ is the query, and $\mathcal B$ acts as the key and value for multi-head attention. Then we incorporate the keyphrase context $\bm{c_t}$ into the decoder hidden state via a feed-forward layer followed by a residual connection (\textsc{RC}):

{\fontsize{10}{11}\selectfont
\vspace{-3mm}
\begin{align}
& \bm h^{d}_t = \textsc{RC}(\bm{W}_{s} \text{tanh} (\bm{W}_{h} \bm h_t + \bm{W}_{c} \bm c_t + \bm b_s), \bm h_t) \text{.} 
\end{align}
\vspace{-3mm}
}

\noindent Finally, the enhanced hidden state $\bm h^{d}_t$ will be passed to another feed-forward layer with softmax to estimate the probability of each output word: 

{\fontsize{10}{11}\selectfont
\vspace{-3mm}
\begin{align}
& P(y_t|\bm{y}_{1:t-1}) = \text{softmax}(\bm W_o \bm h^d_t + \bm b_o)\text{,}
\end{align}
\vspace{-3mm}
}

\noindent where $\bm{W}_*$ and $\bm b_*$ are trainable parameters. 

\subsection{Sentence-level Bag-of-words Planning}
\label{subsec:bow}

We propose an auxiliary task of sentence-level bag-of-words (BOW) planning to supervise the latent representations. The goal is to ground the meaning of the latent representations with the bag-of-words~\cite{fu2020paraphrase} of  target sentences to reflect the global semantic plans. 
Formally, we define the BOW of the $j$-th target sentence $z_j$ as a categorical distribution over the entire vocabulary:

{\fontsize{10}{11}\selectfont
\vspace{-3mm}
\begin{align}
&{p(z_j|\bm {\text{SN}}_j) = \text{softmax}(\text{MLP}(\bm {\text{SN}}_j))\text{,}
}
\label{eq:prior}
\end{align}
\vspace{-3mm}
}

\noindent where $\text{MLP}(*)$ is parameterized as a multi-layer feed-forward network. 
We expect this distribution to capture the overall semantic plan of the corresponding sentence, and enhance $\bm {\text{SN}}$ to guide the surface realization of sentence words by conditioning the probability of each word on the latent representations: $p(y_t|\bm{y}_{1:t-1}, \bm{\text{SN}}_{1:s_{j_t}})$, where $s_{j_t}$ denotes the sentence index of the token $y_t$.
This conditional probability can be naturally satisfied by the autoregressive decoding process. 




The loss of the task is to maximize the likelihood of predicting the BOW of each target sentence:

{\fontsize{10}{11}\selectfont
\begin{align}
& \mathcal{L}_{\text{BOW}} = - \frac{1}{J} \sum_{j}\sum_{l}\log p(z_{jl}|\bm{\text{SN}}_j)
,
\end{align}
}

\noindent where $J$ is the number of target sentence, 
and $p(z_{jl}|\bm{\text{SN}}_j)$ denotes the estimated probability of the $l$-th element in the bag of words for the $j$-th target sentence.


%


\subsection{Coherence-based Contrastive Learning}
\label{subsec:cl}

We further design a contrastive learning (CL)-based training objective to enhance the content planning and drive our model to learn a preference of coherent outputs over incoherent ones. 

\smallskip
\noindent\textbf{Negative Sample Construction.} One challenge for contrastive learning is how to construct negative samples to effectively train the model towards the desired goals. 
We consider the original target as a positive sample representing a logically coherent output with gold planning, and construct negative samples as incoherent ones. In particular, for a positive target, we create 4 negative samples based on the following strategies: (1) \textit{SHUFFLE}, where we randomly shuffle the target sentences to encourage the model to learn the correct sentence order; (2) \textit{REPLACE}, where we randomly replace 50\% of the original target sentences with random sentences from the corpus 
to facilitate the model to learn better content organization; (3) \textit{DIFFERENT}, where we completely replace the original target sentences with a new set that are annotated as the target of a different input from the corpus; 
(4) \textit{MASK}, where we randomly mask 20\% of the non-stop target words that are related to any keyphrases from the keyphrase set, and adopt BART to fill the masked tokens since BART is naturally a denoising model. We enforce the filled negative target to be different from the original one.

\smallskip
\noindent\textbf{Coherence-based Contrastive Loss.}
Since we aim to encourage the model to distinguish between coherent and incoherent targets and generate outputs with coherent logical flows, we design a novel coherence-based contrastive learning objective. Given a source-target pair, the model projects the output feature from the content selection layer to a coherence score between 0 and 1. Formally, for the $i$-th source-target pair, we enforce the score of the original target ($r^+_i$) to be larger than all corresponding negatives ($\{r^-_{ik}\}$) by a fixed margin $\phi$:

{\fontsize{10}{11}\selectfont
\vspace{-3mm}
\begin{align}
& \mathcal{L}_{\text{CL}}(r^+_i, \{r^-_{ik}\}) = \sum_k \text{max}(0, \phi + r^-_{ik} - r^+_i)\text{,} \\
& r^+_i = \text{F}(\text{AvgPool}(\bm W_{cl} \bm H^{d+}_i + \bm b_{cl}))\text{,} \\
& r^-_{ik} = \text{F}(\text{AvgPool}(\bm W_{cl} \bm H^{d-}_{ik} + \bm b_{cl}))\text{,}
\end{align}
\vspace{-3mm}
}

\noindent where $\text{F}(*)$ is a nonlinear transformation with sigmoid, $\bm H^{d+}_i$ and $\bm H^{d-}_{ik}$ are output features from the content selection layer for the positive and the $k$-th negative sample, 
and $\text{AvgPool}(*)$ is the average pooling to compute a fixed-size vector. In this way, we expect the model to assign higher probability to the coherent target than incoherent ones.

\subsection{Training Objective}
We jointly optimize our model for content planning and surface realization by combining the objectives for the sentence-level BOW planning ($\mathcal{L}_{\text{BOW}}$), the word-level generation by cross-entropy loss over the target tokens ($\mathcal{L}_{\text{GEN}}$) , and the contrastive learning loss ($\mathcal{L}_{\text{CL}}$): $\mathcal{L} = \mathcal{L}_{\text{GEN}} + \alpha \mathcal{L}_{\text{BOW}} + \beta \mathcal{L}_{\text{CL}}$, where $\alpha$ and $\beta$ are tuned as hyper-parameters.

%% file: experiments.tex
\begin{table}[t]
\fontsize{9}{12}\selectfont
 \setlength{\tabcolsep}{1.0mm}
  \centering
    \begin{tabular}{ccccccc}
        \toprule
        {\bf Dataset} & {\bf Train} & {\bf Val.} & {\bf Test} & {\bf |State|} & {\bf |Target|} & {\bf \# KP}\\
        \midrule
        {ArgGen} & 42.5k  & 6.5k & 7.5k & 19.4 & 116.6 & 20.6  \\
        {OpinionGen} & 47.6k & 5.0k & 5.0k & 9.0 & 198.2 & 16.2\\
        \bottomrule
    \setlength{\abovecaptionskip}{0.5mm}
    \end{tabular}
    \caption{
    Statistics of the datasets. |State| and |Target| represent number of words of input statement and target, and \#KP denotes the average number of guidance keyphrases.
    }
    \label{tab:data_stat}
    \vspace{-4mm}
\end{table}

\begin{table*}[t]
\fontsize{9}{12}\selectfont
 \setlength{\tabcolsep}{2.2mm}
  \centering
    \begin{tabular}{l cccc c cccc}
        \toprule
        & \multicolumn{4}{c}{\textbf{ArgGen}} & \phantom{} & \multicolumn{4}{c}{\textbf{OpinionGen}} \\
        \cmidrule{2-5} \cmidrule{7-10}
        \textbf{System} & \textbf{BLEU-2} & \textbf{ROUGE-2} & \textbf{METEOR} &
        \textbf{Len.}  &
        \phantom{} &
        \textbf{BLEU-2} & \textbf{ROUGE-2} & \textbf{METEOR} & 
        \textbf{Len.}  \\
        \midrule
        
        \textsc{Retrieval} & 10.95   &  4.02 & 20.70  & 113 & \phantom{} & 18.16  &  6.98 & 24.87  & 153 \\

        \textsc{HierPlan} & 14.29  &  8.38 & 19.03 & 115 & \phantom{} & 10.66    &  5.84  & 17.50 &  107 \\
        
        \textsc{FullSeq2seq} &  36.69    &  26.73 & 42.54 & 97 & \phantom{} &  34.71  &  22.75  & 39.48 &  146\\
        
        \textsc{SepPlan} &  32.38    & 24.84  & 39.79 & 85 & \phantom{} &  31.20   &  19.36  & 33.29 &  151\\
        
        \textsc{SSPlaner} &  36.92    &  26.82  & 42.72 & 105 & \phantom{} &  35.04  &  22.55  & 39.50 &  140 \\
        
        \hdashline
        
        $\textsc{\text{PLANET}}_{\text{w/o CL}}$ &  38.39   &  28.24*  & 44.22* & 99  & \phantom{} &36.41  & \textbf{23.82}*  & 40.84* &  145 \\
        
        \quad $-$ SEL. &  37.66  &  27.71  & 43.76  & 96 & \phantom{} & 35.91  &  23.38  & 40.33 &  142 \\
        
        \quad $-$ BOW &  37.90  &  27.80  & 43.83 & 95 & \phantom{} & 35.68    &  23.42  & 40.39 &  143\\
        
        \textsc{PLANET} (ours) &  \textbf{38.55}*  &  \textbf{28.38}*  & \textbf{44.36}* & 100 & \phantom{} & \textbf{36.79}*   &  23.65*  & \textbf{40.91}* &  146 \\
        \bottomrule
    \end{tabular}
    \vspace{2mm}
    \caption{
    Experimental results on argument generation (ArgGen) and opinion article generation (OpinionGen).  $\textsc{\text{PLANET}}_{\text{w/o CL}}$ is our model variant without contrastive loss.
    We report BLEU-2, ROUGE-2 recall, METEOR and average output lengths (Len.).
   *: significantly better than all other methods without asterisks (Welch’s t-test, $p<0.05$).
  }
  \vspace{-3mm}
  \label{tab:auto-eval}
\end{table*}

\subsection{Tasks and Datasets}
We conduct experiments on two long-form opinion generation datasets of distinct domains: (1) Argument Generation (\textbf{ArgGen})~\cite{hua-etal-2019-argument-generation}, where the model is required to generate a counter-argument to refute a given proposition; (2) Opinion Article Generation (\textbf{OpinionGen})~\cite{hua-wang-2020-pair}, to produce an opinion article given a title. The data statistics are shown in Table~\ref{tab:data_stat}.

\smallskip
\noindent\textbf{Argument Generation.} We first apply data from Reddit \textit{r/ChangeMyView} (CMV) for argument generation. We consider the original poster (OP) title as the statement, and the high-quality argument replies (with community endorsement) as the targets.
Note that we consider the full argument replies as targets. The noun phrases and verb phrases that contain at least one topic signature word~\cite{lin-hovy-2000-automated} are extracted to form the guidance keyphrases.

\smallskip
\noindent\textbf{Opinion Article Generation.} For generating opinion articles, we consider samples from the New
York Times (NYT) corpus~\cite{sandhaus2008new}, with articles  whose taxonomy labels include \textit{Top/Opinion}. 
The articles with less than three sentences or more than 10 sentences are discarded. We further exclude articles containing more than 250 tokens considering the limited computing resources. 57,600 articles are randomly selected as the final dataset. 
We apply the same method as in argument generation to extract topical guidance keyphrases. The article title is regarded as the input statement.

\subsection{Baselines and Comparisons}
We compare our model against the following baselines
: (1) \textbf{\textsc{Retrieval}}~\cite{stab-etal-2018-argumentext} which retrieves targets based on TF-IDF weights of words from the training set. We keep the top-ranked results as outputs; (2) \textbf{\textsc{HierPlan}}~\cite{hua-etal-2019-argument-generation} which is an end-to-end trained generation model with a hierarchical decoder to perform sentence-level content planning and surface generation; 
(3) \textbf{\textsc{FullSeq2seq}}~\cite{schiller-etal-2021-aspect} where we fine-tune BART with keyphrases concatenated to the input statements; 
(4) \textbf{\textsc{SSPlaner}}~\cite{kang-hovy-2020-plan} is a global planning method which first conducts content prediction and then guides the surface generation with the predicted contents;
(5) \textbf{\textsc{SepPlan}} is a two-stage planning model similar to ~\citet{hua-wang-2020-pair}, where we first fine-tune a BART as the planner to generate the ordered keyphrase plans for each target sentence, and then fine-tune another BART as the generator to produce final outputs based on the statement and keyphrase plans. The details of \textsc{SepPlan} are in the Appendix~\ref{subsec:training_decoding_details}.

\subsection{Training and Decoding Details}
We use the BART-base version in all experiments for both our method and baselines.
We truncate both input statement and output target to at most 256 tokens during training. 
For the BOW planning loss ($\mathcal{L}_{\text{BOW}}$), we consider the salient content words as the ground-truth bag of words for each target sentence.
For the training objective, we set $\alpha$ as 0.2 for ArgGen and 0.3 for OpinionGen, and $\beta$ as 0.2 based on the validation performance. The margin for contrastive loss is set as 0.5 for ArgGen and OpinionGen according to the validation performance. We optimize our model with AdamW~\cite{loshchilov2017decoupled}. 
During the decoding time, we apply nucleus sampling~\cite{holtzman2019curious} with a cumulative probability threshold of 0.9, and the maximum of generation steps are 150 for ArgGen and 200 OpinionGen.
More training and decoding details are in the Appendix~\ref{subsec:training_decoding_details}.

%% file: results.tex
\subsection{Automatic Results}
We first evaluate our model with BLEU~\cite{papineni-etal-2002-bleu}, ROUGE~\cite{lin-2004-rouge}, and METEOR~\cite{denkowski-lavie-2014-meteor}. 
The results are shown in Table~\ref{tab:auto-eval}.

Our $\textsc{\text{PLANET}}_{\text{w/o CL}}$ model (without contrastive loss) consistently outperforms all baseline methods. In particular, compared with \textsc{FullSeq2seq} and \textsc{SSPlaner} which are also fine-tuned based on BART with the same inputs, the substantial improvements underscore the effectiveness of our dynamic content planning to generate better outputs. Meanwhile, the significant lead over \textsc{HierPlan} indicates the importance of incorporating content planning into pre-trained language models. 
Furthermore, $\textsc{\text{PLANET}}_{\text{w/o CL}}$  significantly outperforms \textsc{SepPlan}, which confirms that the end-to-end training in our approach can mitigate the disconnection issue of the two-stage generation pipeline and produce superior results. 

Among our model variants, removing content selection (w/o SEL.) and BOW planning (w/o BOW) both lead to performance decrease. This demonstrates the importance of the components that help the model conduct effective content planning. In addition, we observe that incorporating the contrastive loss (\textsc{\text{PLANET}}) brings performance gains on automatic results, especially with significant improvements on BLEU scores. This suggests that \textit{our contrastive loss can guide the model to more precisely use keyphrases and reflect the keyphrase information in the outputs}. We provide further analysis on the keyphrase usage in Section \ref{subsec:human}.

\begin{figure}[t]
    \centering
    \includegraphics[scale=0.73]{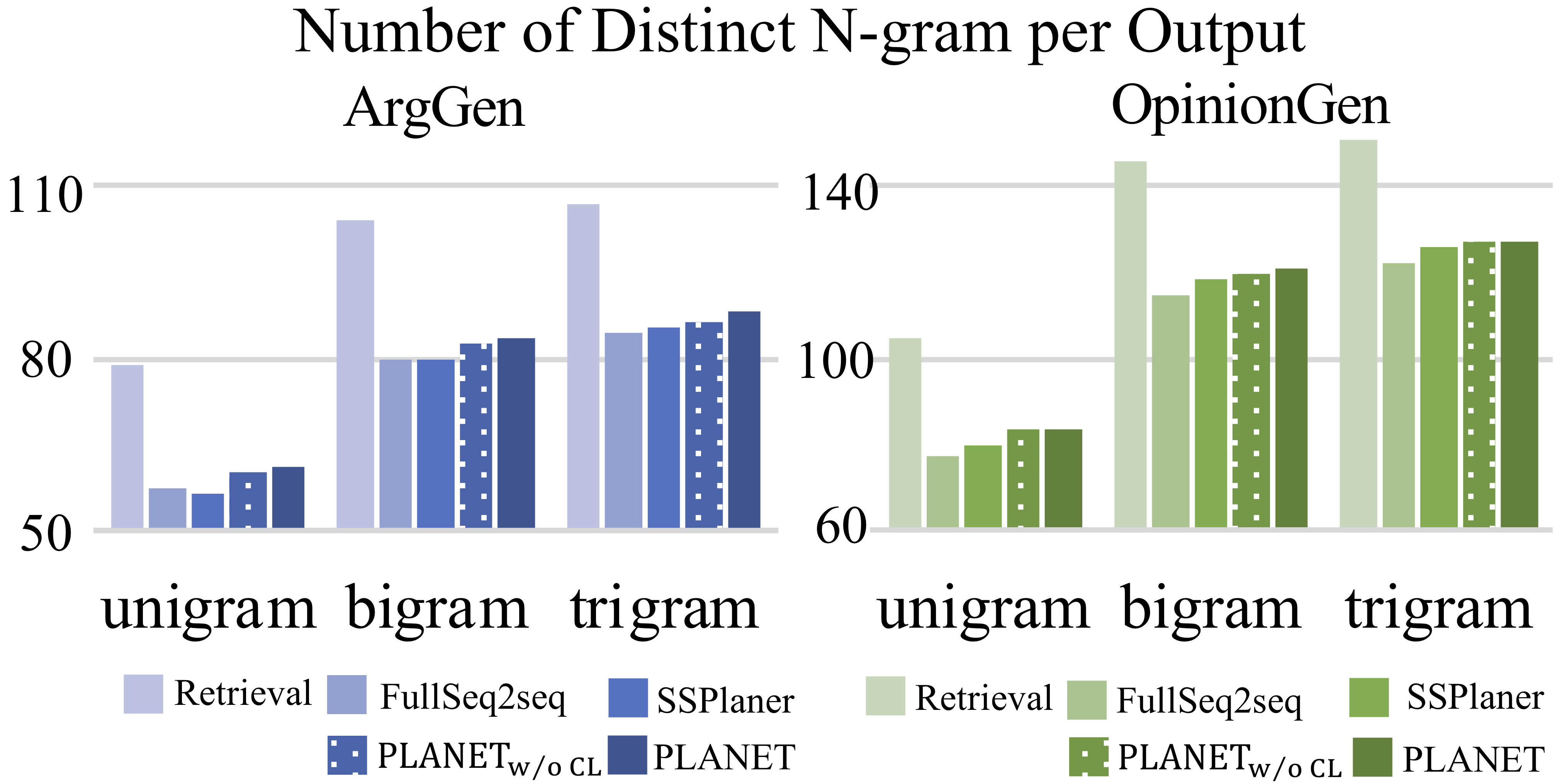}
    \vspace{-1mm}
    \captionof{figure}{ 
    Average number of distinct n-grams per output.
    }
    \label{fig:dist-eval}
    \vspace{-2mm}
\end{figure}

\smallskip
\noindent \textbf{Content Richness.} 
To evaluate content richness, we employ Distinct $n$-gram ~\cite{Li2016naacl} that calculates the number of distinct $n$-grams per output in Figure~\ref{fig:dist-eval}. \textsc{Retrieval} achieves the highest distinct results on both datasets since it returns top-ranked human-written texts with the most distinct words. Among generative methods, our dynamic planning model $\textsc{\text{PLANET}}_{\text{w/o CL}}$ outperforms all baselines on both datasets. In addition, after applying contrastive loss, our \textsc{\text{PLANET}} model generates even more unique $n$-grams. The results imply our dynamic content planning and contrastive loss can enable the model to generate richer contents. 

\begin{figure}[t]
    \centering
    \includegraphics[scale=0.28]{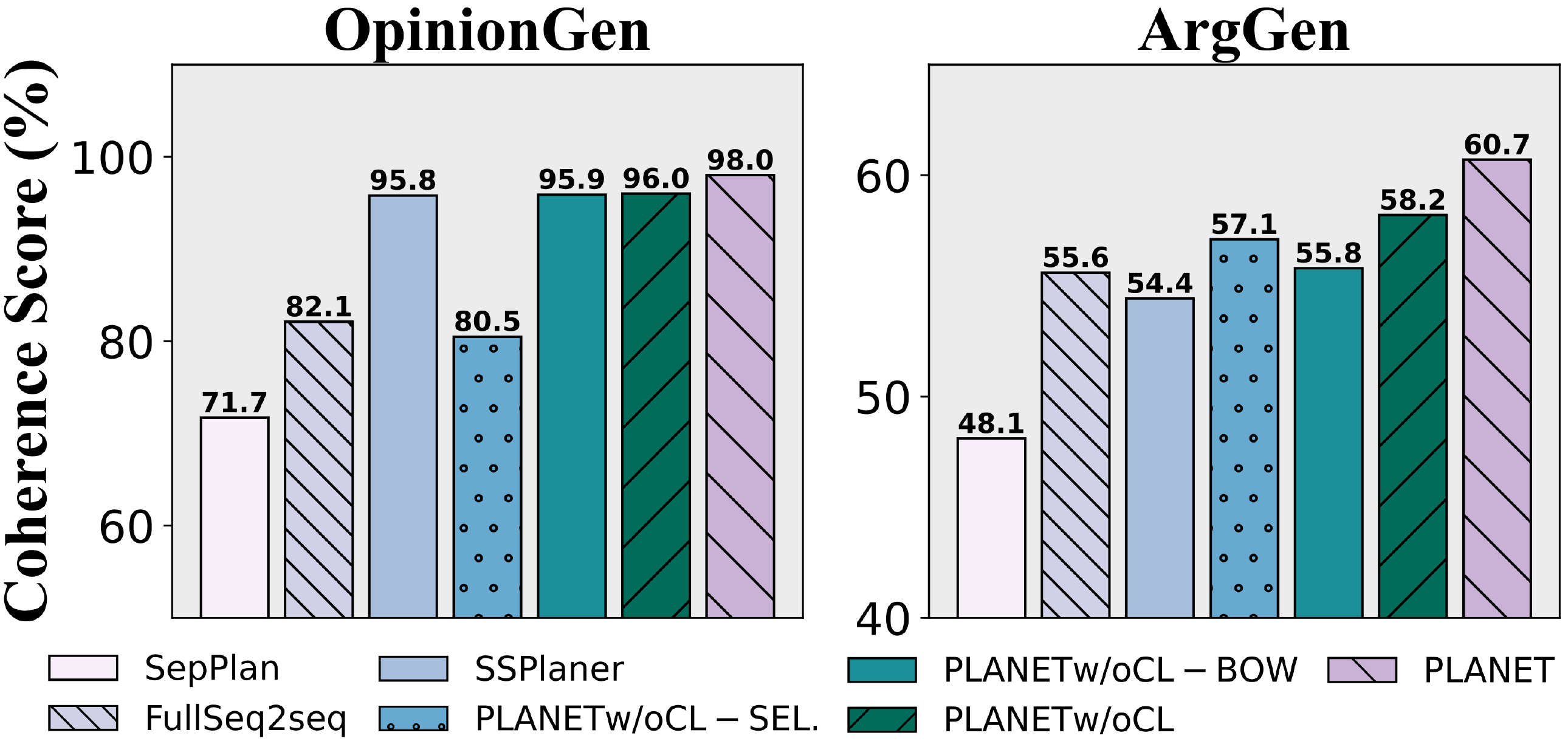}
    \vspace{1mm}
    \captionof{figure}{ 
    Automatic evaluation on output coherence.
    }
    \label{fig:coherence-eval}
    \vspace{-2mm}
\end{figure}

\smallskip
\noindent \textbf{Automatic Evaluation on Coherence.} 
We fine-tune BERT~\cite{devlin-etal-2019-bert} on each dataset to automatically evaluate the output coherence, which predicts a score between 0 and 1 for each output. The higher score indicates a more coherent output. The coherence model details are in Appendix~\ref{subsec:coherence_model}.


The results are shown in Figure~\ref{fig:coherence-eval}. Among all methods, \textsc{\text{PLANET}} achieves the highest coherence scores on both datasets, suggesting that our dynamic planning and contrastive loss are effective to improve the coherence of  outputs. In contrast, \textsc{SepPlan} has the lowest scores, indicating that decoupling planning and decoding stages may lead to cascading errors.
Compared to \textsc{FullSeq2seq} and \textsc{SSPlaner}, our $\textsc{\text{PLANET}}_{\text{w/o CL}}$ model without contrastive loss also maintains better coherence, which confirms that incorporating dynamic content planning essentially promotes coherence for long text generation. Moreover, we observe that the results on OpinionGen are consistently better than those on the ArgGen dataset. A possible reason is that arguments in ArgGen are collected from social networks and contain more colloquial and informal expressions, making it harder to learn the implicit logical coherence. We leave this for future work.

\begin{table}[t]
\fontsize{9}{12}\selectfont
 \setlength{\tabcolsep}{1.8mm}
  \centering
    \begin{tabular}{lcc}
        \toprule
        {\bf System} & {\bf OpinionGen (\%)} & {\bf ArgGen (\%)}  \\
        \midrule
        \textsc{PLANET} &  98.03 & 60.71\\
        \quad w/o {SHUFFLE}  & 96.20  & 59.30 \\
        \quad w/o {REPLACE}  & 96.02  & 58.41 \\
        \quad w/o {DIFFERENT}  &  96.11 & 59.95 \\
        \quad w/o {MASK}  & 96.16  & 59.58\\
        \bottomrule
    \end{tabular}
    \vspace{2mm}
    \caption{
    Coherence scores for different negative strategies.
  }
  \label{tab:cl_ablation}
  \vspace{-5mm}
\end{table}

\smallskip
\noindent\textbf{Ablation on Contrastive Sample Construction.}
We study the contribution of each negative sample construction strategy for improving the coherence of the outputs. As in Table~\ref{tab:cl_ablation}, removing each strategy leads to a performance degradation, indicating the effectiveness of all types of negative samples to enhance the contrastive learning. Among all negatives, removing \textit{REPLACE} shows the most effects on both datasets. We hypothesize that replacing target sentences breaks the original logical flow and thus is more likely to encourage the model to focus on the global coherence. In contrast, \textit{DIFFERENT} shows the least effects. One possible explanation is that this strategy focuses more on topical relatedness between the input and output, instead of the logical flow within the output as the negative sample itself is inherently coherent.


\subsection{Human Evaluation}
\label{subsec:human}
We hire three proficient English speakers as human judges to evaluate model outputs on a scale of 1 (worst) to 5 (best) for: (1) \textbf{topic relatedness} which measures whether the output is relevant
and consistent to the input; (2) \textbf{coherence} which measures the high-level logical flow and transition among sentences; and (3) \textbf{content richness}, measuring the amount of informative talking points and specific details. We also ask judges to select top-ranked results based on the overall quality, and ties are allowed. 50 random samples are selected from each task.
The detailed guidelines of human evaluations are provided in the Appendix~\ref{sec:human_eval_guideline}.

The results are shown in Table~\ref{tab:human_eval_res}. Both our model variants achieve better results than \textsc{FullSeq2seq} on all aspects, underscoring the effectiveness of our dynamic planning to promote output coherence. Moreover, introducing contrastive objective further improves output quality on the above aspects, and the outputs are more likely to be top-ranked. Overall, the human results verify the capability of our dynamic planning and contrastive objective to generate high-quality long-form texts.

\begin{table}[t]
\fontsize{8}{12}\selectfont
 \setlength{\tabcolsep}{1.5mm}
  \centering
    \begin{tabular}{llcc lc}
        \toprule 
        {\bf Task} & {\bf Model} & {\bf Rel.} & {\bf Coh.} & {\bf Rich.} & {\bf Top-1} \\
        \midrule
        ArgGen & \textsc{FullSeq2seq}  &   2.25  & 2.47 & 2.57 & 20.7\%  \\
        &  $\textsc{\text{PLANET}}_{\text{w/o CL}}$ & 2.79  & 2.83 & 3.10 & 30.0\% \\
        & \textsc{PLANET}  & \bf 2.83  & \bf 2.89 & \bf 3.21 & \bf 33.3\% \\
        \midrule
        OpinionGen & \textsc{FullSeq2seq} & 3.65 & 3.19 & 3.44 & 16.0\% \\
        & $\textsc{\text{PLANET}}_{\text{w/o CL}}$   & 3.81 & 3.27 & 3.64  & 28.7\%\\
        & \textsc{PLANET} & \bf 3.89  & \bf 3.47  & \bf 3.81 & \bf 37.3\%\\
        \bottomrule
    \end{tabular}
    \vspace{2mm}
    \caption{
    Human evaluation on relatedness (Rel.), coherence (Coh.), content richness (Rich.) and \% of evaluations a model being ranked in top 1 based on the overall quality. All Krippendorff’s $\alpha \geq 0.34$, with specific values in the Appendix~\ref{sec:human_eval_guideline}.
  }
  \label{tab:human_eval_res}
  \vspace{-2mm}
\end{table}

\begin{figure}[t]
    \centering
    \includegraphics[scale=0.35]{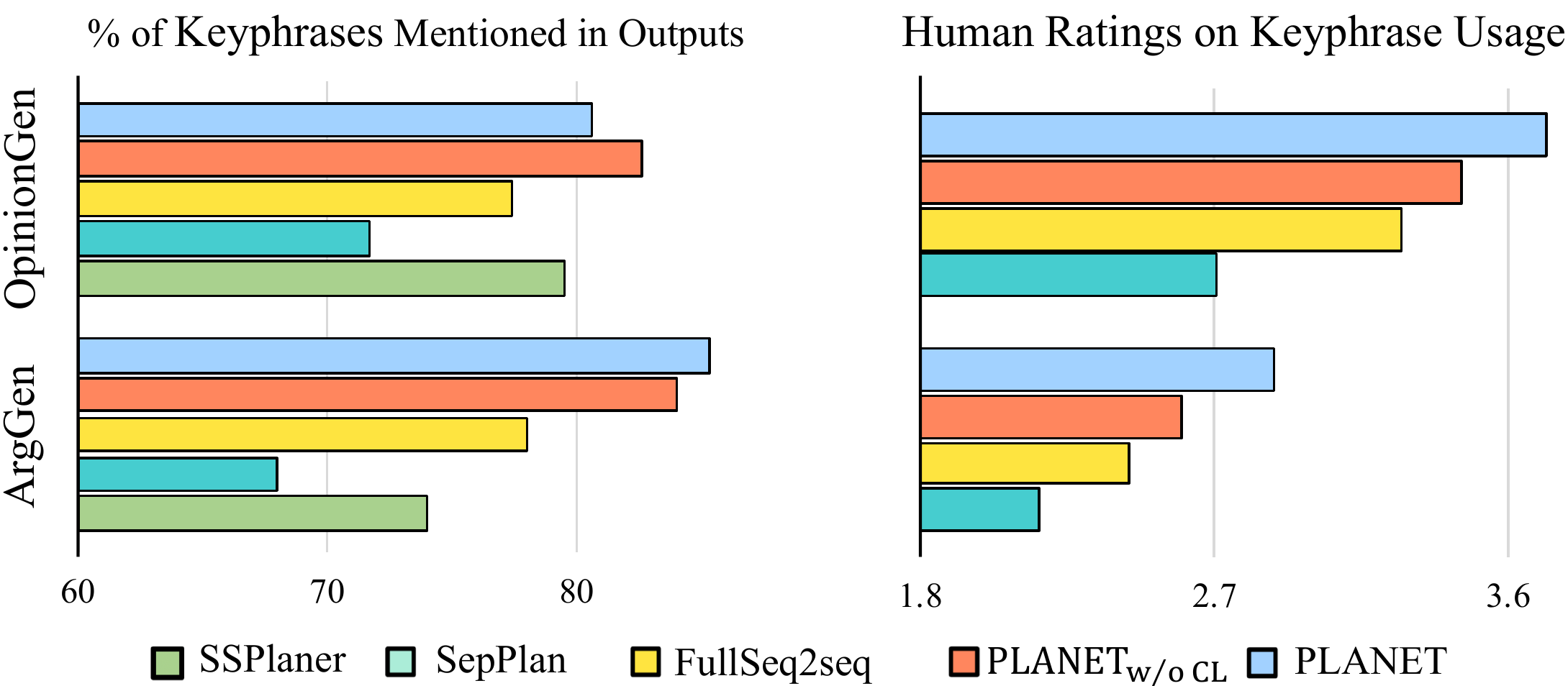}
    \vspace{-3mm}
    \captionof{figure}{ 
    Left: percentage of keyphrases that are mentioned in outputs. Right: human evaluation on keyphrase usage.
    }
    \label{fig:kp_usage}
    \vspace{-2mm}
\end{figure}

\begin{figure}[t]
    \def\arraystretch{1.5}
	\fontsize{9}{10}\selectfont
	\setlength{\tabcolsep}{0.8mm}
	\centering
	\begin{tabular}
	{|p{76mm}|}
    \hline
	\textbf{Statement}: Absolute free speech doesn`t exist - real debate is about where to draw the line \\
	\hline
	\textsc{\textbf{PLANET}}:  {\scriptsize \circled{1}} Absolute \underline{free speech} exists in a {\color{red!70}\underline{\textbf{conceptual}}}, {\color{blue!75}\textbf{legal}}, and {\color{green!70!blue}\textbf{practical}} way. {\scriptsize \circled{2}} The {\color{purple!70}\textbf{\underline{simple} fact}} that it exists is \underline{obvious}.  {\scriptsize \circled{3}} If you're going to say something in {\color{blue!75}\textbf{a \underline{legal} way}}, at any time, the {\color{teal}\textbf{\underline{consequences}}} should be the consequences.  {\scriptsize \circled{4}} It's not that absolute \underline{free} speech doesn't exist, but the {\color{red!70}\textbf{conceptual way}} you could interpret it is much {\color{green!70!blue}\textbf{more \underline{practical}}}. {\scriptsize \circled{5}} {\color{orange!65}\textbf{The \underline{internet}}} has made it much more obvious that you could use it any way you wanted.
	\\
	\begin{tabular}{p{1mm}p{73mm}}
     &
     \textbf{BOW}:  
	{\scriptsize \circled{1}} free, speech, {concept} 
	{\scriptsize \circled{2}} {simple}, obvious {\scriptsize \circled{3}} {consequences}, {legal}, {illegal} 
	{\scriptsize \circled{4}} freedom, case, {practical} 
	{\scriptsize \circled{5}} {internet}, easy
	\end{tabular}
	\\
	\hdashline
	\textsc{\textbf{FullSeq2seq}}: Absolute free speech exists in a conceptual (and probably legal) way. There is the simple fact that absolute free speech exists in a legal way. It's obvious what the consequences are for someone, but it can be done any time. In many cases, it's very practical to do something like this in a legal way because the internet makes it very obvious.
	\\
    
    \hline\hline
	\textbf{Statement}: Millions of mothers can't choose not to work \\
	\hline
	\textbf{\textsc{PLANET}}: {\scriptsize \circled{1}} {\color{red!70}\textbf{Single \underline{mothers}}} have to decide how to raise their \underline{children}. {\scriptsize \circled{2}} But the \underline{choice} \underline{mothers} have has often been made to \underline{work}, {\color{teal}\textbf{juggling financial responsibilities}} that make it all the more difficult. {\scriptsize \circled{3}} While it is true that many {\color{orange!65}\textbf{\underline{families}}} have {\color{magenta!70}\textbf{no such \underline{choice}}}, the {\color{green!70!blue}\textbf{reality is the same}}: {\color{red!70}\textbf{single \underline{mothers}}} have {\color{magenta!70}\textbf{little options}}. {\scriptsize \circled{4}} And while it is true that {\color{orange!65}\textbf{a \underline{family} of two or more}} {\color{blue!75}\textbf{lowers \underline{pay}}} and is likely to take many times the {\color{purple!70}\textbf{cost of similar \underline{work}}}, the reality is that it may not be that simple (...)
	\\
	\begin{tabular}{p{1mm}p{73mm}}
     &
	\textbf{BOW}: 
	{\scriptsize \circled{1}} child, parent, children 
	{\scriptsize \circled{2}} work, choice, mother
	{\scriptsize \circled{3}} choice, family, mother 
	{\scriptsize \circled{4}} work, pay, children, family\\
	\end{tabular}
	\\
	\hdashline
 	{\textbf{\textsc{FullSeq2seq}}}: Crittenden is right about single mothers' choice to choose not to work, in her book "the choice mothers make" But the sad reality of working families is that it is the reality that Ms. Crittenden and many others, in juggling financial responsibilities, are forced to choose not to work. If they are lucky enough to be able to keep their jobs, they can be at similar work as nannies. But the sad reality is that the choice mothers make is no longer one wage earner (...)
	\\
    \hline
	\end{tabular}
\caption{Sample outputs on ArgGen (Upper) and OpinionGen (Lower). For our model results, the phrases relevant to the guidance keyphrases are highlighted in colors, and the words related to the corresponding  BOW are underlined. Best viewed in color. 
} 
\label{fig:samples}
\vspace{-3mm}
\end{figure}

\smallskip
\noindent\textbf{Appropriateness of Keyphrase Usage.}
We further study how keyphrases are utilized in outputs. 
We first compute the percentage of keyphrases mentioned in outputs, as in
the left of Figure~\ref{fig:kp_usage}. Among all models, \textsc{SepPlan} uses the least keyphrases in final outputs. However, its intermediate planning results cover more than 95\% of keyphrases. This confirms that the two-stage method results in a disconnection problem between the planning module and the surface realization module, and the outputs are not guaranteed to reflect the plans. Compared to \textsc{FullSeq2seq} and \textsc{SSPlaner}, our methods cover more keyphrases, suggesting that our dynamic planning and keyphrase selection are useful to help the model better incorporate the guidance signal into outputs.

We further select 50 random samples for both tasks and ask the same human judges to score the outputs from 1 (worst) to 5 (best) on the correctness of keyphrase usage: whether the model uses keyphrases adequately as main talking points when generating outputs. Results in Figure~\ref{fig:kp_usage} (right) indicate that our models tend to use more keyphrases and properly organize them in the outputs compared to all baseline methods. Although on OpinionGen our contrastive model  mentions fewer keyphrases, human judges rate it with higher scores for keyphrase usage.
We speculate that this can be attribute to the \textit{MASK} strategy for negative sample construction in contrastive learning, which helps to improve the model ability on the appropriate usage of keyphrases. The above results confirm that \textsc{PLANET} can properly utilize the keyphrases and reflect the contents in the outputs.

\subsection{Sample Outputs and Discussions}
We show two sample outputs on both tasks and highlight the phrases relevant to the guidance keyphrases in Figure~\ref{fig:samples}. We can see that on both tasks, our model effectively leverages guidance keyphrases as main talking points, and properly organizes and reuses the keyphrases to form a coherent output. In contrast, \textsc{FullSeq2seq} suffers from incoherence issues such as repetition (e.g., the first and second argument sentences) and inconsistent stance (e.g., ``\textit{choose not to work}'' in generated opinion article). This indicates that our dynamic planning is effective to guide the model to better leverage keyphrases in the outputs.

We also present the predicted BOW of our model for each generated sentence. As can be seen, our model predicts most of the salient content words of the target sentences and effectively reflects the semantic plans in the generated sentences, suggesting that our latent representations are useful to capture the global semantic information of each sentence and conduct content planning during the generation process. However, there is still a large gap
compared with human written texts, inspiring the future work on long-form text generation.
More sample outputs are provided in Appendix~\ref{sec:additional_samples}.

%% file: conclusion.tex
We present a novel generation framework to dynamically conduct content planning and surface realization in large autoregressive Transformers by leveraging self-attention and high-level latent representations. 
The latent representations are grounded by bag-of-words that measures the overall semantic plan of each target sentence.
We further introduce a novel coherence-based contrastive objective with different negative sample construction strategies to improve output coherence. Experiment results on two opinion text generation tasks demonstrate that our model can generate high-quality outputs with better coherence and content richness. 

%% file: ethics.tex
We recognize that our method may generate fabricated and potentially harmful contents due to the systematic biases of pre-training using heterogeneous web corpora and the open-ended generation characteristics of the opinion generation tasks. Therefore, we urge the users to carefully examine the ethical influence of the generated outputs and cautiously apply the system in real-world applications. 

%% file: supplementary/details.tex


\subsection{Additional Experimental Results}
In table~\ref{tab:auto-eval} we report automatic results on both tasks. Here we present additional automatic results of BLEU-3 and ROUGLE-L (recall) in Table~\ref{tab:add_cmv} and Table~\ref{tab:add_nyt}.

\subsection{Training and Decoding Details}
\label{subsec:training_decoding_details}
\noindent\textbf{Model Training.}
Our model is built based on BART, and we use $\text{BART}$-base version for all experiments. Our model contains 185M parameters in total. 
The batch size is set to be 8, and the maximum training epoch is set as 15 for non-contrastive training and 18 for contrastive training.
We truncate both the input statement and output target to be at most 256 tokens during training. We resize the BART embedding matrix with a new token [SN] and insert a [SN] token before each target sentence. This is also done for baselines for a fair comparison.
For computing resources, we use NVIDIA Tesla V100 GPUs with 32 GB memory for all experiments, and utilize the mixed-precision (FP16) to improve the computational efficiency. 
For contrastive learning, for each positive target, we construct 4 negatives using the strategies described in Section~\ref{subsec:cl} respectively. The best model checkpoint is chosen based on the validation loss. 
Our model takes around 4-5 hours for training, and 30 minutes for decoding on V100 GPUs. 


\smallskip
\noindent\textbf{Decoding.}
During decoding time, we apply the nucleus sampling~\cite{holtzman2019curious}, and set $k=10$ and $p=0.9$. Considering the computational cost, we limit the maximum of generation steps to 150 for argument generation on ArgGen and 200 for opinion article generation on OpinionGen.
To reduce variance introduced by sampling-based decoding method, we decode three times and average the results for automatic evaluations.
For our model, we enforce each target sentence to start with a [SN] token during inference: we pre-define a list of sentence end markers, and when the model finishes generating a sentence, we enforce the next generated token to be [SN], although we find in most cases the model can automatically generate [SN]. The generation process stops when the model generates the <EOS> token. In this way, the model can automatically decide on how many sentences to be generated, and conduct content planning and surface realization in a dynamic way.


\begin{table}[t]
\fontsize{9}{11}\selectfont
 \setlength{\tabcolsep}{1.8mm}
  \centering
    \begin{tabular}{lcc}
        \toprule
        {\bf System} & {\bf BLEU-3 (\%)} & {\bf ROUGE-L (\%)}  \\
        \midrule
        \textsc{Retrieval} & 4.52 & 16.13 \\
        \textsc{HierPlan} & 9.28 & 19.11\\
        \textsc{FullSeq2seq} & 25.83 & 26.88 \\
        \textsc{SepPlan} & 22.17 & 23.24 \\
        \textsc{SSPlaner} & 25.85 & 26.99 \\
        \hdashline
        \textsc{PLANET} &  27.11 & 27.42* \\
        \quad $-$ SEL.  & 26.58 & 27.01 \\
        \quad $-$ BOW  & 26.78 & 26.97 \\
        \textsc{PLANET} (ours)  &  \bf{27.21}* & \bf{27.54}* \\
        \bottomrule
    \end{tabular}
    \vspace{2mm}
    \caption{
    Additional experimental results of BLEU-3 and ROUGE-L (recall) on ArgGen.
  }
  \label{tab:add_cmv}
  \vspace{-3mm}
\end{table}

\begin{table}[t]
\fontsize{9}{11}\selectfont
 \setlength{\tabcolsep}{1.8mm}
  \centering
    \begin{tabular}{lcc}
        \toprule
        {\bf System} & {\bf BLEU-3 (\%)} & {\bf ROUGE-L (\%)}  \\
        \midrule
        \textsc{Retrieval} & 10.98 & 17.99 \\
        \textsc{HierPlan} & 5.81 & 15.98\\
        \textsc{FullSeq2seq} & 25.71 & 26.29 \\
        \textsc{SepPlan} &  21.23 & 21.68 \\
        \textsc{SSPlaner} & 25.67 & 26.49 \\
        \hdashline
        \textsc{PLANET} &  26.91 & 27.08* \\
        \quad $-$ SEL.  & 26.49 & 26.79 \\
        \quad $-$ BOW  & 26.40 & 26.77 \\
        \textsc{PLANET} (ours)  &  \bf{27.01}* & \bf{27.18}* \\
        \bottomrule
    \end{tabular}
    \vspace{2mm}
    \caption{
    Additional experimental results of BLEU-3 and ROUGE-L (recall) on OpinionGen.
  }
  \label{tab:add_nyt}
  \vspace{-3mm}
\end{table}

\smallskip
\noindent\textbf{Evaluation Scripts.} We use NLTK~\footnote{https://www.nltk.org/} to implement BLEU and METEOR, and the ROUGE\_SCORE package~\footnote{https://pypi.org/project/rouge-score/} to implement ROUGE.

\smallskip
\noindent\textbf{Details for \textsc{SepPlan}.}
We design a two-stage generation method, \textsc{SepPlan}, as a baseline model by fine-tuning two independent BART models for content planning and surface realization respectively, similar to ~\citet{hua-wang-2020-pair}. In particular, the planner BART takes a statement and unordered keyphrase as inputs, and autoregressively generates content plans as a sequence of tokens for every target sentence, where each content plan is represented by the ordered keyphrases with the same order as they appear in the corresponding sentence. Segmenter is added between sentence plans to indicate the sentence boundary. Then the generator BART consumes the concatenation of the statement and content plans to produce the final results. During training, the ground-truth content plans are used to train the generator, and during inference the predicted plans are used. For decoding, we apply beam search for the planner and nucleus sampling for the generator. Note that ~\citet{hua-wang-2020-pair} applies BERT as planner in their original paper, and we replace BERT with BART as BART gives better performance in our experiments.

\subsection{Training Details for Coherence Model}
\label{subsec:coherence_model}
We propose a neural coherence model to evaluate output coherence. Concretely, we fine-tune BERT~\cite{devlin-etal-2019-bert} on each dataset to compute the coherence scores. 
Instead of computing the overall coherence scores by
measuring and aggregating the coherence of its adjacent sentence pairs~\cite{xu-etal-2019-cross}, we fine-tune BERT on the whole text to better learn the global coherence~\cite{xing-carenini-2021-improving}.

For training, we follow ~\citet{sharma-etal-2019-entity} and adopt hinge loss to teach the model to assign higher scores to coherent targets than incoherent ones. The score is normalized into [0, 1] with  $\text{sigmoid}$ function, and the margin is set to be 0.8. Since each target usually contains multiple sentences, we insert a separator token [SEP] between each adjacent sentence pair. For data construction, we consider the original text as a positive sample, and randomly shuffle sentences to construct negative ones. The test accuracy  is 94.3\% on OpinionGen and 73.0\% on ArgGen, respectively. This implies that our coherence model can be used as a reliable metric to evaluate the output coherence.

%% file: supplementary/human_eval.tex
We present 55 random samples on each task for human evaluation, and the first 5 samples are used only for calibration~\footnote{The payment for each human judge is 20 dollars per hour.}. We anonymize the models and shuffle the outputs to the annotators. We evaluate model outputs on the following aspects, and the detailed guidelines are in Table~\ref{tab:human_eval}:

\begin{itemize}[noitemsep,nolistsep,wide]
    \item {\textbf{Relatedness}}: 
    whether the output is relevant and consistent to the input;
    \item {\bf Coherence:} whether the overall logical flow is appropriate and the transitions among sentences are natural and smooth;
    \item {\bf Content Richness:} whether outputs contain substantial talking points and convey specific details;
    \item {\bf Overall Ranking:} this is a general assessment that whether you think the output ranks top among all candidates. Ties are allowed, which means you can choose multiple outputs as top-ranking for a sample.
\end{itemize}

\begin{table}[t]
\fontsize{9}{12}\selectfont
 \setlength{\tabcolsep}{1.8mm}
  \centering
    \begin{tabular}{lcccc}
        \toprule
        {\bf Task} & {\bf Rel.} & {\bf Coh.} & {\bf Rich.} & {\bf KP-Use.}  \\
        \midrule
        ArgGen & 0.49 & 0.34 &  0.40 & 0.44 \\
        OpinionGen & 0.41 & 0.46 & 0.37 & 0.36 \\
        \bottomrule
    \end{tabular}
    \vspace{2mm}
    \caption{
    Krippendorff’s $\alpha$ for human evaluation on relatedness (Rel.), coherence (Coh.), content richness (Rich.) and keyphrase usage (KP-Use.).
  }
  \label{tab:agreement}
  \vspace{-7mm}
\end{table}

To measure agreement among human judges, we compute Krippendorff’s $\alpha$ for each aspects. The values for all aspects on both datasets are presented in Table~\ref{tab:agreement}. As can be seen, all values are equal or larger than 0.34, indicating a general consensus among the judges.

%% file: supplementary/limitations.tex
Here we discuss the limitations of our work and the potential directions for future studies. Long-form text generation is a challenging task which requires the model to properly select and organize contents, and faithfully reflect the plans in surface realization, in order to form a coherent output. The results suggest that our dynamic content planning can effectively leverage keyphrases and generate more coherent and richer texts than strong baseline methods. Nevertheless, there is still a gap compared with human written outputs. Also, in this paper we follow previous 
work to study the keyphrases guided generation~\cite{hua-wang-2020-pair,rashkin-etal-2020-plotmachines}, where we assume the availability of keyphrases as guidance signals. For the scenarios where guided keyphrases are not available in test time, one can use either retrieval-based methods~\cite{hua-etal-2019-argument-generation,wu2020controllable} or a separate knowledge-enhanced generative module to obtain guided keyphrases. However, this is out of the scope of this work.

We believe there are several promising directions to explore in the future. First direction can be applying our dynamic planning method into pretrainning or post-pretrainning stage. One advantage of our model is that it does not require additional annotated data (the keyphrases and BOW labels can be automatically constructed with off-the-shelf tools as described in data processing). Leveraging massive pretraining data would be very helpful to further improve the model performance on long-text generation in various domains.

Second, one can study different supervision signals to train the latent representations. In this work we apply bag-of-words to ground the latent representations, which aims to capture the overall semantic information. Other supervision signals such as discourse structures and entity usage are also very important for modeling coherence. Considering these aspects into planning can further improve the output coherence. Meanwhile, coherence is a broad definition including topical relatedness, 
causal relationship, temporal ordering and discourse structures~\cite{li-jurafsky-2017-neural}. Designing different supervision signals to tackle specific aspects for coherence would also be a promising direction.

Third, in this work we consider keyphrases as guidance signals to control the generation. Future work can incorporate different guidance signals from heterogeneous sources such as structured knowledge and commonsense information to further improve the output quality.

%% file: supplementary/samples.tex
We present additional examples on argument generation in ArgGen and opinion article generation in OpinionGen from Figure~\ref{fig:samples1} to Figure~\ref{fig:samples4}.

\begin{table*}
	\fontsize{10}{12}\selectfont
	 \def\arraystretch{1.8}
    \centering
    \begin{tabular}
    {lp{120mm}}
         \toprule
         \multicolumn{2}{c}{\textbf{Relatedness}} \\
         \midrule
         \rowcolor{lightgray!30}
          1 & The output is very generic and irrelevant to the statement  \\
          
          3 & The output is tangential to the statement and mentions some relevant concepts or entities, but in general is not precisely on topic \\
          
          \rowcolor{lightgray!30}
          5 &  The output is highly relevant to the statement and fully on topic \\

          \midrule
         \multicolumn{2}{c}{\textbf{Coherence}} \\
         \midrule
         \rowcolor{lightgray!30}
         1 & The sentences are completely disconnected and unrelated, without any overall logical structure \\

         3 & The output sentences are somehow connected or on the same topic, but the transition is not smooth, and the overall logical flow is not strong to convey a progression of idea \\

         \rowcolor{lightgray!30}
         5 & The output is completely coherent with natural and smooth transition among sentences, and the high-level logical flow is strong to form a good progression \\

          \midrule
         \multicolumn{2}{c}{\textbf{Content Richness}} \\
         \midrule
         \rowcolor{lightgray!30}
         1 & The output is very generic with no specific information or talking points \\

         3 & The output contains several talking points or key information to form an opinion text \\

         \rowcolor{lightgray!30}
         5 & The output contains sufficient key points and informative contents to from a strong counter-argument or opinion article. \\
         
         \bottomrule
         \vspace{-1mm}
    \end{tabular}
    \caption{Explanations on human evaluation for aspect scales.}
    \label{tab:human_eval}
\end{table*}

\begin{figure*}[t]
    \bgroup
    \def\arraystretch{1.8}
	\fontsize{10}{12}\selectfont
     \hspace{-1mm}
	\setlength{\tabcolsep}{0.8mm}
	\centering
	\begin{tabular}{p{150mm}}
	\toprule
	\textbf{Statement}: CMV: journalism has become so biased in this country that it has lost its relevance. \\
	\midrule
	\textsc{\textbf{PLANET}}:  {\footnotesize \circled{1}} True \text{{journalism}} is \textbf{biased by definition}. {\footnotesize \circled{2}} You have to have \textbf{sources} in order to read the news, especially when \textbf{the facts and propaganda} are coming out. {\footnotesize \circled{3}} You need to know \textbf{a piece} or two about their policies. {\footnotesize \circled{4}} This includes reading from \textbf{a wide range of opinions}. {\footnotesize \circled{5}} You can be \textbf{swayed by the emotion} of rage. {\footnotesize \circled{6}} A lot of \textbf{great journalists} and pundits are coming out with their favor. {\footnotesize \circled{7}} The AP is no actor. {\footnotesize \circled{8}} BBC has been relentless in \textbf{their editorial and their propaganda}. {\footnotesize \circled{9}} NPR has tried to \textbf{push conservative candidates} and push their favor in a public and this is reflected in their \textbf{editorial and news}.
	\\
	\begin{tabular}{p{2mm}p{145mm}}
     &
	\textbf{BOW}:  
	{\footnotesize \circled{1}} journal, biased, media
	{\footnotesize \circled{2}} news, fact, propaganda, media
    {\footnotesize \circled{3}} news, media, 
    {\footnotesize \circled{4}} people, political, lot
    {\footnotesize \circled{5}} emotion, rage, news
    {\footnotesize \circled{6}} good, public, news,
    {\footnotesize \circled{7}} media, main
    {\footnotesize \circled{8}} BBC, news, 
    {\footnotesize \circled{9}} public, candidates, political, popular
	\end{tabular}
	\\
	\hdashline
	\textsc{\textbf{FullSeq2seq}}: The AP is biased by definition. They are biased because they are biased by their bias. A lot of true journalism is achieved by the emotion of outrage and propaganda, and no actor can objectively prove this. They are biased by their biases, because they want to push conservative candidates' policies and favorites to an audience that is more informed than they are themselves. So if you want a piece of true journalism from a public that doesn't want to read the news, then you need sources and sources to get the facts. You need great journalists to listen to your sources. BBC, NPR, the AP, and the like.
	\\
    \bottomrule
	\end{tabular}
    \vspace{-3mm}
\caption{Sample outputs on ArgGen. For our results, we present the bag-of-words predicted for each sentence, and the phrases relevant to the guidance keyphrases are highlighted in bold.
} 
\label{fig:samples1}
\egroup
\vspace{-4mm}
\end{figure*}

\begin{figure*}[t]
    \bgroup
    \def\arraystretch{1.8}
	\fontsize{10}{12}\selectfont
     \hspace{-1mm}
	\setlength{\tabcolsep}{0.8mm}
	\centering
	\begin{tabular}{p{150mm}}
	\toprule
	\textbf{Statement}: CMV: the European Parliament should be elected on transnational lists \\
	\midrule
	
	\textsc{\textbf{PLANET}}:  {\footnotesize \circled{1}}  A big issue with the EU is that \textbf{the formation of government} would \textbf{face a similar problem} as the \textbf{federalization} of the United States. {\footnotesize \circled{2}}  \textbf{The US constitution} included \textbf{major guarantees} for autonomy, a \textbf{separate national voice} for all and a broader level of \textbf{governance}. {\footnotesize \circled{3}}  A lot of \textbf{Euro-skeptic countries} have been concerned about secession at the start of the EU, but the EU has been able to \textbf{reduce these concern}. {\footnotesize \circled{4}} Euro-skeptic countries are more likely to \textbf{balance state autonomy}, and thus need a \textbf{slow-motion version} of government. {\footnotesize \circled{5}} if the EU were to step down, it would be violating by the
	\\
	\begin{tabular}{p{2mm}p{145mm}}
     &
	\textbf{BOW}:  
	{\footnotesize \circled{1}} government, issue, EU, federalization
    {\footnotesize \circled{2}} democracy, constitution, governance, independence
    {\footnotesize \circled{3}} EU, countries, states
    {\footnotesize \circled{4}} government, federal, state
	{\footnotesize \circled{5}} violated, EU, constitution, government
	\end{tabular}
	\\
	\hdashline
	\textsc{\textbf{FullSeq2seq}}: The United States was a major hot point in US politics from the start, and was a slow-motion version of the US constitution that included major guarantees of independence, and a separate national voice for all. Many countries now want to balance state autonomy with autonomy. This would mean that many would face a similar problem with the federalization of the US. But it would also reduce these concern to a broader level. I don't think we should be at war with the EU for the federalization of the US, but that is a big issue. It would be very different from creeping federalization of the US as a whole. I don't think that the United States would do well without a big issue like the
	\\
    \bottomrule
	\end{tabular}
    \vspace{-3mm}
\caption{Sample outputs on ArgGen. For our results, we present the bag-of-words predicted for each sentence, and the phrases relevant to the guidance keyphrases are highlighted in bold.
} 
\label{fig:samples2}
\egroup
\vspace{-4mm}
\end{figure*}

\begin{figure*}[t]
    \bgroup
    \def\arraystretch{1.8}
	\fontsize{10}{12}\selectfont
     \hspace{-1mm}
	\setlength{\tabcolsep}{0.8mm}
	\centering
	\begin{tabular}{p{150mm}}
	\toprule
	\textbf{Statement}: Endangered species will gain from reform \\
	\midrule
	\textsc{\textbf{PLANET}}: {\footnotesize \circled{1}} To the editor: I agree with your April 19 \textbf{column attack legislation} to reform \textbf{the endangered species act}. {\footnotesize \circled{2}} But it is important to focus on how it \textbf{improves the recovery} of endangered species. {\footnotesize \circled{3}} In an April 28 editorial, \textbf{representative Richard Pombo} of California and 125 other democrats called on the federal government to \textbf{provide habitat} for the endangered species. {\footnotesize \circled{4}} This would have been good for the species as well as the economy. {\footnotesize \circled{5}} It would have been good for the species, as it would help the economy. {\footnotesize \circled{6}} The endangered species act allows homeowners to \textbf{protect habitat} without \textbf{forcing private landowners} to pay taxes to protect it. {\footnotesize \circled{7}} If landowners were allowed to sell off their \textbf{critical habitat}, they would pay for the maintenance of homes and homes. {\footnotesize \circled{8}} And if owners paid taxes to protect habitat, it would help \textbf{attract these species} to their area. {\footnotesize \circled{9}} David h. Lindman  Washington, April 30, 1997 the writer is an environmental advocate.
	\\
	\begin{tabular}{p{2mm}p{145mm}}
     &
	\textbf{BOW}:  
	{\footnotesize \circled{1}} endangered, act, April, species
	{\footnotesize \circled{2}} species, endangered, protect, 
    {\footnotesize \circled{3}} protect, endangered, species, land, support
	{\footnotesize \circled{4}} life, species, economy,
	{\footnotesize \circled{5}} protect, species, park
	{\footnotesize \circled{6}} owners, species, protection, bill
    {\footnotesize \circled{7}} landowners, habitat, park, housing, 
	{\footnotesize \circled{8}} homes, build, land, 
    {\footnotesize \circled{9}} writer, address
	\end{tabular}
	\\
	\hdashline
	\textsc{\textbf{FullSeq2seq}}: To the editor: while i applaud your call to reform the endangered species act, your April 19 column attack legislation that would allow the states to force private landowners to provide habitat for endangered species. In an April 28 editorial, representative Richard Pombo of Texas and 125 other democrats wrote that ``the species cannot be exploited to attract these species to this program.'' However, there are other ways to exploit these species: the endangered species act is a law requiring the state to provide habitat for endangered species and requiring the states to provide a plan to protect habitat for the species. If the endangered species act is enacted, it will be in effect, and will be a significant step toward conservation. The bill is a response to the plight of the endangered species act and will help improve its financing. Daniel s. Bennett New York, April 30, 1999 the writer is chairman of the house appropriations committee.
	\\
    \bottomrule
	\end{tabular}
    \vspace{-3mm}
\caption{Sample outputs on OpinionGen. For our results, we present the bag-of-words predicted for each sentence, and the phrases relevant to the guidance keyphrases are highlighted in bold.
} 
\label{fig:samples3}
\egroup
\vspace{-4mm}
\end{figure*}

\begin{figure*}[t]
    \bgroup
    \def\arraystretch{1.8}
	\fontsize{10}{12}\selectfont
     \hspace{-1mm}
	\setlength{\tabcolsep}{0.8mm}
	\centering
	\begin{tabular}{p{150mm}}
	\toprule
	\textbf{Statement}: Protect inmates' rights \\
	\midrule
	\textsc{\textbf{PLANET}}: {\footnotesize \circled{1}} To the editor: \textbf{Elisabeth Sara} (op-ed, Jan. 1) argues that \textbf{stripping inmates} of \textbf{civil legal assistance} would not result in a lasting peace or in freeing them from incarceration. {\footnotesize \circled{2}} Yet the Bush administration has not spent the last 10 years \textbf{exorcising the demons} of discrimination against federally \textbf{financed prisoners}. {\footnotesize \circled{3}} Prisoners' \textbf{legitimate civil rights complaints} should be heard in court. {\footnotesize \circled{4}} it is time for the administration to begin \textbf{stripping inmates} of their rights and \textbf{redeeming the debts} owed them by the government. {\footnotesize \circled{5}} The government must make sure that they receive \textbf{adequate funds} for housing, education and legal services. {\footnotesize \circled{6}} It must ensure that their rights are protected and that they receive the same level of support for the \textbf{legal services} they deserve. {\footnotesize \circled{7}} David Jacobs  New York, Jan. 2, 2002 the writer is a fellow at the Brennan center for justice, Columbia university.
	\\
	\begin{tabular}{p{2mm}p{145mm}}
     &
	\textbf{BOW}:  
	{\footnotesize \circled{1}} editor, prison, rights, civil
	{\footnotesize \circled{2}} prisoners, legal, treatment, fair
	{\footnotesize \circled{3}} legitimate, rights, prison
	{\footnotesize \circled{4}} system, stripping, access
    {\footnotesize \circled{5}} housing, cost, services, affordable
    {\footnotesize \circled{6}} support, money, services, deserve
    {\footnotesize \circled{7}} writer, director, center, new, york
	\end{tabular}
	\\
	\hdashline
	\textsc{\textbf{FullSeq2seq}}: To the editor: Elisabeth Sara's article ``exorcising the demons of jail'' (April 19) is a powerful reminder that the system of state-financed prisoners' rights is stripping inmates of their right to equal opportunity, access to legal services and a chance to make real progress in redeeming the debts they have been stripped to the bone. Without adequate financing and affordable legal counsel, thousands of prison-financed prisoners will never be able to get their civil legal assistance they deserve. Joseph Jacobs  new York, April 20, 2006 the writer is a staff attorney at the Brennan center for justice.
	\\
    \bottomrule
	\end{tabular}
    \vspace{-3mm}
\caption{Sample outputs on OpinionGen. For our results, we present the bag-of-words predicted for each sentence, and the phrases relevant to the guidance keyphrases are highlighted in bold.
} 
\label{fig:samples4}
\egroup
\vspace{-4mm}
\end{figure*}